\documentclass[]{bytedance_seed}

\usepackage[toc,page,header]{appendix}
\usepackage{tablefootnote}
\usepackage{amsmath}
\usepackage{amsfonts}
\usepackage{amssymb}
\usepackage{graphicx}
\usepackage{booktabs} 
\usepackage{array}    
\usepackage{pdflscape} 
\usepackage{cleveref}
\usepackage{pdfpages}

\crefname{table}{Table}{Tables} 

\usepackage{minitoc}
\usepackage{colortbl}
\usepackage{xcolor}
\usepackage{amssymb}

\definecolor{keywordcolor}{rgb}{0.7, 0.1, 0.1}   
\definecolor{tacticcolor}{rgb}{0.0, 0.1, 0.6}    
\definecolor{commentcolor}{rgb}{0.4, 0.4, 0.4}   
\definecolor{symbolcolor}{rgb}{0.0, 0.1, 0.6}    
\definecolor{sortcolor}{rgb}{0.1, 0.5, 0.1}      
\definecolor{attributecolor}{rgb}{0.7, 0.1, 0.1} 

\title{Seed-Prover: Deep and Broad Reasoning for\\ Automated Theorem Proving}

\author{ByteDance Seed AI4Math}

\abstract{
LLMs have demonstrated strong mathematical reasoning abilities by leveraging reinforcement learning with long chain-of-thought, yet they continue to struggle with theorem proving due to the lack of clear supervision signals when solely using natural language. 
Dedicated domain-specific languages like Lean provide clear supervision via formal verification of proofs, enabling effective training through reinforcement learning. 
In this work, we propose \textbf{Seed-Prover}, a lemma-style whole-proof reasoning model. Seed-Prover can iteratively refine its proof based on Lean feedback, proved lemmas, and self-summarization.
To solve IMO-level contest problems, we design three test-time inference strategies that enable both deep and broad reasoning. Seed-Prover proves $78.1\%$ of formalized past IMO problems, saturates MiniF2F, and achieves over 50\% on PutnamBench, outperforming the previous state-of-the-art by a large margin.
To address the lack of geometry support in Lean, we introduce a geometry reasoning engine \textbf{Seed-Geometry}, which outperforms previous formal geometry engines.
We use these two systems to participate in IMO 2025 and fully prove 5 out of 6 problems.
This work represents a significant advancement in automated mathematical reasoning, demonstrating the effectiveness of formal verification with long chain-of-thought reasoning.
}

\checkdata[Project Page]{\url{https://github.com/ByteDance-Seed/Seed-Prover}}

\begin{document}
\maketitle
\vspace{-20pt}

\begin{figure}[h]
	\centering
        \includegraphics[width=0.84\linewidth]{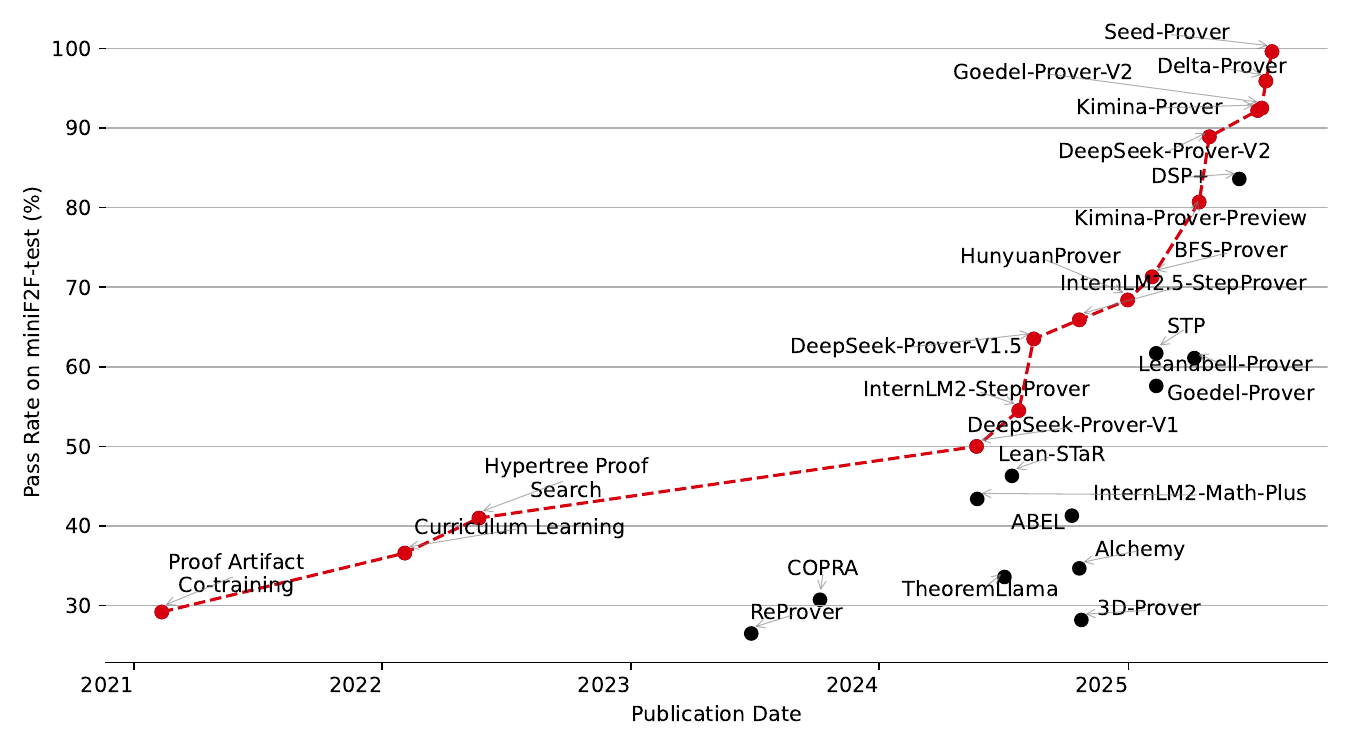}
		\caption{Growth in MiniF2F-Test performance over time.}
		\label{fig:miniF2F_SOTA}
	\vspace{-0.5in}
\end{figure}

\section{Introduction}

Recent advances in large language models (LLMs) have shown extending reasoning lengths through natural language can significantly boost performance on math benchmarks such as MATH and AIME \citep{openai2024openaio1card,deepseekai2025deepseekr1incentivizingreasoningcapability}. 
Training such models requires reinforcement learning (RL) on verifiable answers. 
It is extremely difficult to automatically, or even manually, verify a proof in natural language since each step must be carefully checked for correctness \citep{petrov2025proofbluffevaluatingllms}.
This poses significant challenges for applying reinforcement learning to the training of large language models to prove mathematical statements.
Unlike natural language, formal languages such as Lean can provide a clear and automatic signal on the correctness of a formalized proof.    
A noteworthy work from AlphaProof \citep{alphaproof} uses Lean to successfully solve 3 problems from the 2024 International Mathematical Olympaid (IMO).
AlphaProof demonstrates that LLMs using formal language are capable of proving very challenging problems that LLMs using natural language fail to prove.

There are two types of LLM formal provers, step-level provers \citep{alphaproof,polu2022formal,wu2024LeangithubcompilinggithubLean,wu2024internlm25stepproveradvancingautomatedtheorem,xin2025bfsproverscalablebestfirsttree} and whole-proof generation provers\citep{xin2025deepseekproverv,dong2025stpselfplayllmtheorem}.
Step-level provers incrementally generate Lean code line-by-line. While this enables close interaction with the Lean environment, it requires special scaffolding to generate a complete Lean proof, and the interaction is often too granular to allow high-level reasoning.
In contrast, whole-proof models generate an entire Lean proof at once, but typically lack interaction with the Lean compiler.
Recent work has shown that combining whole-proof models with long chain-of-thought reasoning \citep{kimina,deepseek-v2,lin2025goedel} substantially outperforms step-level provers.
In this work, we propose \textbf{Seed-Prover}, a whole-proof model with following features:
\begin{itemize}
    \item \textbf{Lemma-Style Proving}: Seed-Prover tries to generate useful intermediate lemmas before proving the main statement. These lemmas serve as shared knowledge across different inference paths.
    \item \textbf{Iterative Proof Refinement}: Seed-Prover iteratively refines its proof based on Lean compiler feedback, previous proved lemmas, and self-summarization.
    \item \textbf{Test-Time Scaling}: We implement a three-tiered inference strategy that enables Seed-Prover to think both deeply and broadly—allocating thinking budget to fine details while exploring interesting properties.
    \item \textbf{SOTA Performance}: Seed-Prover proves 5 out of 6 problems in IMO 2025, saturates MiniF2F \citep{zhengminif2f} (shown in Figure~\ref{fig:miniF2F_SOTA}), and outperforms prior work by up to $3\times$ on multiple formal benchmarks.
\end{itemize}

Due to the lack of sufficient geometry support in Lean, Seed-Prover incorporates a dedicated geometry reasoning engine \textbf{Seed-Geometry}. Similar to existing line of efforts in AlphaGeometry \cite{trinh2024solving,chervonyi2025gold} and TongGeometry \cite{zhang2024proposing}, Seed-Geometry follows the forward-chaining design in the reasoning engine implementation, where the system derives all known facts by checking applicable rules until closure is reached. By backward-tracing fact dependencies, Seed-Geometry identifies the minimum dependency relations in a geometry problem's configuration, seperating the problem context from the auxiliary constructions necessary to prove a problem. Using statistics derived from more than past 20 years of math olympiad competitions, Seed-Geometry performs extensive search in the geometry space defined by its dedicated domain-specific language and establishes a repository of 230 million unique geometry problems requiring auxiliaries. A Seed model trained on such dedicated geometry data becomes an exceptionally effective neuro-symbolic geometry prover, where it fills in the missing auxiliary geometry elements and the geometry reasoning engine performs step-by-step forward-chaining, completing the final proof of a problem. In experiments, Seed-Geometry solves 43 of the IMO-AG-50 (\textit{vs.} 42 by AlphaGeometry 2), a benchmark that curates geometry problems of IMO from 2000 to 2024. It also sets a new state-of-the-art on the IMO shortlist geometry problems from 2000 to 2022, and notably solves the geometry problem of IMO 2025 under just 2 seconds.
\section{Approach}
Here we introduce the two systems we used in IMO 2025, Seed-Geometry and Seed-Prover.

\subsection{Seed-Geometry}
Seed-Geometry builds on the success of TongGeometry \cite{zhang2024proposing} and performs a major redesign. From a global perspective, both systems leverage trained neural models to complete missing auxiliary constructions and specialized reasoning engines to forward-chain derivation. However, Seed-Geometry is a substantial upgrade over TongGeometry in the following aspects.

\subsubsection{Extended Domain-Specific Language} 
Seed-Geometry constructs geometric diagrams in the principle of ruler-and-compass construction. However, plain ruler-and-compass construction steps can be long and cumbersome, making the language representation of the constructions overly verbose, introducing unnecessary burden on both the neural processing of Transformer-based large language model and the symbolic derivation of the backend engine. To mitigate these issues, Seed-Geometry groups particular action sequences into specific actions, making the representation of the problem concise enough. Of particular note, Seed-Geometry has several such composite actions: isogonal conjugate with respect to a triangle and a point, exsimilitude center of two circles, insimilitude center of two circles. All three actions can be represented with primitive ruler-and-compass actions; yet the construction sequence in itself is non-trivial and unnecessarily clumsy. 

\subsubsection{Extremely Fast Reasoning Engine}
Seed-Geometry improved its reasoning engine's performance by rewriting its backend in C++ and making it accessible to Python users through Pybind11. This change led to roughly 100-fold speed increase compared to the Python implementation in TongGeometry. The C++ implementation handles memory more efficiently and benefits from compiler optimizations, allowing for much faster \textbf{deep} searches within the reasoning engine. This is particularly crucial because the engine's forward-chaining design typically slows down considerably when the search tree expands widely at deeper levels.

\subsubsection{Exceptional Large Language Model}
Seed-Geometry utilizes a high-performing large language model from the Seed family \cite{seed2025seed1}. This particular Seed model has undergone extensive pre-training on vast datasets of coding and mathematics, granting it a wide array of specialized skills. The specific model size was chosen considering the number of data tokens. We considered training two models in an actor--critic setup initially: the policy model that proposed possible next auxiliary element to construct and a value model that predicted the number of steps to go from the state. However, we note from preliminary experiments that a single Seed model serving as the policy would suffice, contradicting the design of both a policy model and a value model in existing work \cite{zhang2024proposing}. We also note that a policy model unspecialized to the specific goal makes both training and solving more manageable and therefore, we only trained the model on pairs of problem context and auxiliaries, without the target fact goal in the prompt.

\subsubsection{Extensive Search}
When presented with a new problem, Seed-Geometry first transforms the representation into a canonical form. If the reasoning backend successfully finds the goal fact to prove in the reasoning process, the problem is considered immediately solved. Otherwise, Seed-Geometry initiates a search process. In particular, Seed-Geometry employs beam search, with the policy model generating proposals for each beam in the buffer. With the extremely fast reasoning engine, Seed-Geometry supports running each new generated proposal in time. If any one of the proposal leads to the proof, the problem is considered proven; otherwise we select the top few proposals for the next step of expansion in the search tree based on each proposal's cumulative negative log likelihood. The search process terminates until a fixed number of steps have been consumed. Seed-Geometry's search process is made efficient and scalable. Compared to TongGeometry, Seed-Geometry's solving process supports a distributed setup where each GPU process communicates with each other and blocked only at the beam selection point. Each GPU process is also equipped with a CPU thread pool that asynchronously executes the reasoning step during language model inference such that the reasoning cost can be overlapped with language model inference. 

\subsection{Seed-Prover}
Seed-Prover is a large language model specialized in formal reasoning in Lean 4. Its most significant distinction from prior work lies in its adoption of lemma-style proving as the proof paradigm, which places \textbf{lemmas} at the center of the reasoning process. This approach offers several key advantages: it enables clear identification of lemmas that have and have not been proved, indicating the progress made in solving the main problem; lemmas can be processed independently and combined freely; lemmas from different inference trajectories can be combined to address more challenging problems. Both the training and inference procedures of Seed-Prover are designed around lemmas.

\subsubsection{Lemma-Style Proving}
\begin{figure}[t]
	\centering
        \includegraphics[width=0.9\linewidth]{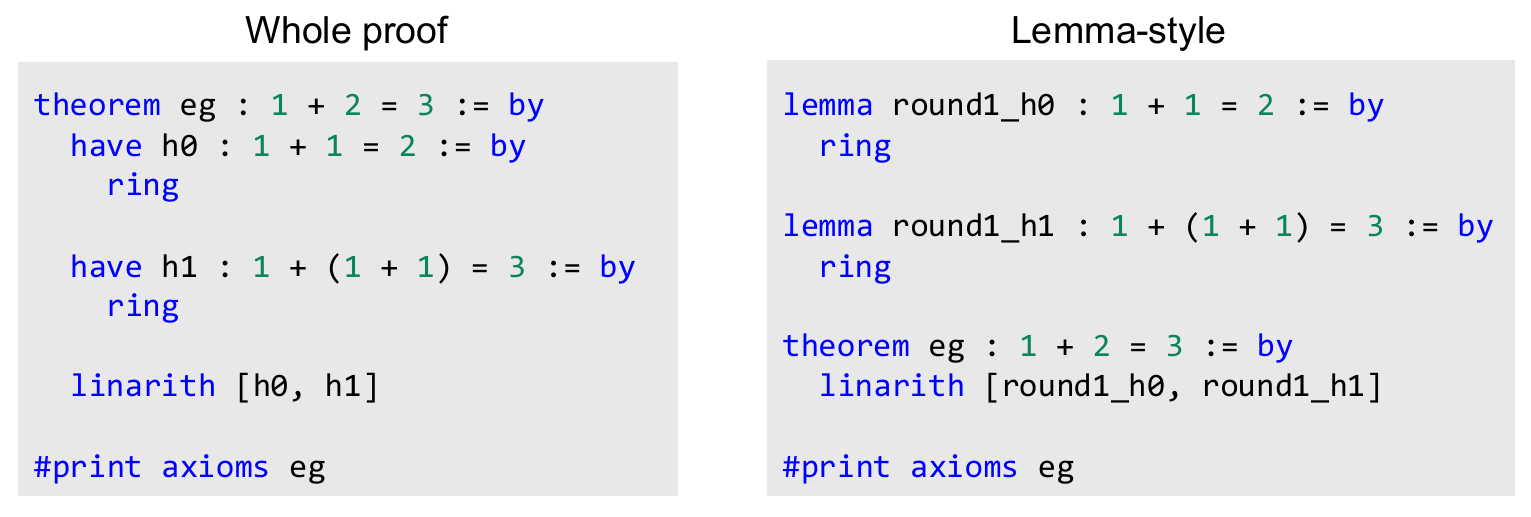}
		\caption{An example of whole proof and lemma-style proof in Lean 4.}
		\label{fig:lemma-style}
	\vspace{-0.15in}
\end{figure}

Previous works \citep{kimina, deepseek-v2, lin2025goedel} trained the model to generate whole proofs starting with the keyword \textit{theorem}. In contrast, we first require the model to generate some useful lemmas---each introduced by the keyword \textit{lemma}---before generating the main proof using \textit{theorem} by applying the generated lemmas.
An example is shown in Figure~\ref{fig:lemma-style}.
This lemma-style proof provides following merits.
First, it allows clear identification of the lemmas that have been successfully proved, and those that need further refinement.
Second, lemmas are modular---they can be compiled independently, stored independently, and combined freely.
Additionally, proofs of lemmas may provide inspiration to the model for proving unproved lemmas and the main problem.
To enable this workflow, we establish a lemma pool for each difficult problem, which stores comprehensive data from all our inference runs, including lemma statements, lemma names, complete proofs, proof difficulties, and dependency relations.
The lemma pool is typically used to (1) retrieve the most relevant lemmas by name or formal statement; (2) sample the most difficult lemmas according to their proof difficulty.

\subsubsection{Conjecture Proposing}
When tackling challenging contest-level math problems, human contestants often identify interesting properties of the problem and use them to guide their reasoning. Seed-Prover is trained to propose such potentially useful properties by chain-of-thought reasoning. Notably, this process differs from chain-of-thought reasoning to solve the problem directly; rather, it emphasizes broad exploration of the problem space without committing to a particular approach. Take functional equations as an example: one might conjecture that the function is \textit{injective}, \textit{surjective}, \textit{bijective}, \textit{monotonic}, or \textit{periodic}—all without engaging in deep, targeted reasoning about the problem itself. In a sense, many useful properties of a problem may be enumerated before full problem resolution.
The proposer module accepts an unsolved problem and, optionally, some already proved lemmas as input, and generates 10–50 candidate conjectures about properties of the problem. Proposing multiple conjectures in parallel significantly increases diversity and the likelihood of covering the valuable properties.
For each problem, we may repeat this process multiple times to create a large conjecture pool.

This approach differs from Draft, Sketch, Prove \citep{jiangdraft}, which presumes the ability to fully solve the problem upfront. In contrast, our method performs broad exploratory searches over the problem space enabling discovery of useful properties, which makes it possible to solve problems that the model cannot solve directly in natural language.

\subsubsection{Training}
To enable seamless interaction between Seed-Prover and Lean, we adopt multi-stage, multi-task reinforcement learning (RL) based on VAPO \citep{vapo}. The RL reward is 1 if the formal statement is successfully proven, and 0 otherwise. Additionally, a formatting penalty is applied to encourage the model to generate lemmas before attempting the main theorem. As training progresses, problem diffculty, problem quality, and maximum output length are progressively increased. The training dataset comprises a combination of open-source datasets \citep{yinglean, numina_math_datasets, albalak2025bigmathlargescalehighqualitymath, yu2025dapo, peng2025criticlean} and in-house formalized problems. For problems that are too difficult for single-pass generation, we use our proposer to generate easier problem variants and put these into the training dataset. We also exclude problems which are too easy (i.e. proof rate above 1/4) from RL training.

Unlike prior work \citep{kimina, deepseek-v2} that only utilize formal statements as prompts for RL training, our approach randomly incorporates natural language hints, natural language proofs, similar lemmas, proved lemmas, failed lemmas, failed attempts, summaries of previous attempts, and Lean compiler feedback into the prompt. This diverse prompting strategy enhances the model's adaptability within our inference pipeline by enabling it to understand and utilize various types of input.

\subsubsection{Test-Time Scaling}

\begin{figure}[t]
	\centering
        \includegraphics[width=0.9\linewidth,page=1]{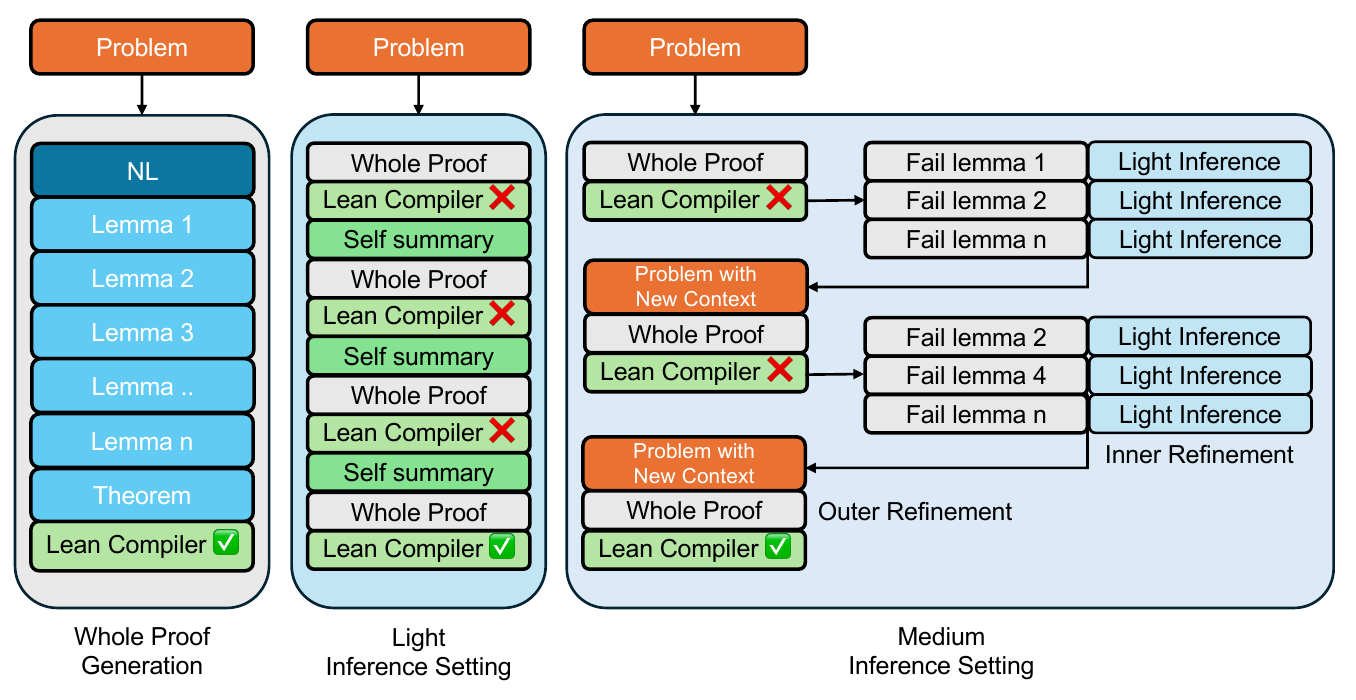}
		\caption{The workflows of single-pass whole proof generation, light, and medium inference settings.}
		\label{fig:test-time-1}
	\vspace{-0.15in}
\end{figure}

\begin{figure}[t]
	\centering
        \includegraphics[width=0.9\linewidth,page=2]{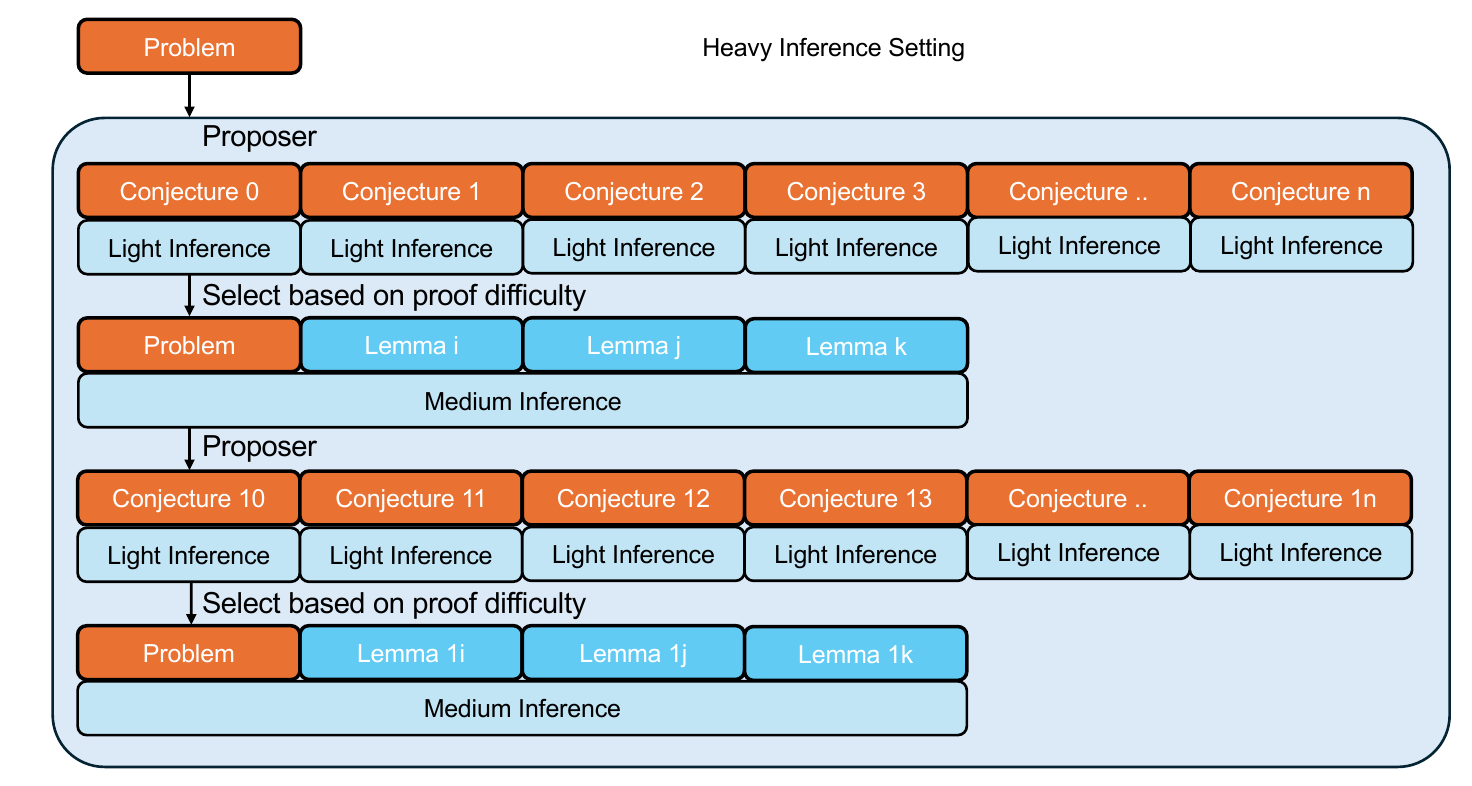}
		\caption{The workflow of heavy inference setting.}
		\label{fig:test-time-2}
	\vspace{-0.15in}
\end{figure}

Here, we introduce our approach to test-time scaling of Seed-Prover. Depending on available inference budgets and problem difficulties, we developed three levels of strategies, which are illustrated in Figure~\ref{fig:test-time-1} and Figure~\ref{fig:test-time-2}.

\paragraph{Light} Previous work evaluated LLM theorem provers by Pass@$k$. We find that iteratively refining proofs using Lean compiler feedback \citep{deltaprover} in conjunction with self-summarization can surpass the limits of single-pass inference token budgets and improve proof ability significantly. In the light setting, each proof attempt is refined up to 8–16 times and evaluated under Pass@8–16. We denote the sample budget of Pass@$n$ and up to $m$ refinements as $n \times m$, so the sample budget of the light setting is equivalent to generating the whole proof at Pass@64–256. This setting completes in 1–2 hours. Under this setting, Seed-Prover proves IMO 2022 P2 ($\textit{MOHS}=15$\footnote{MOHS is Math Olympiad Hardness Scale: \url{https://web.evanchen.cc/upload/MOHS-hardness.pdf}. We note that the hardness scale for human contestants may not be well-aligned with the difficulty of proving it in Lean using an LLM.}), whereas without refinement, the same problem can only be proved in Pass@8192.
We observed two dominant behaviors of Seed-Prover in the light inference setting. 
First, it fixes Lean syntax errors in response to Lean compiler feedback.
Second, it refines initial proof sketches, a process that might entirely alter the reasoning trajectory.

\paragraph{Medium}
Proofs for challenging competition problems are often lengthy and structurally complex. The medium test-time setting introduces both outer and inner refinement processes. The outer refinement process mirrors the light setting which refines the proof of the original main problem. The inner refinement process targets difficult lemmas that the outer refinement process generates but fails to prove, using a light setting with an $8 \times 8$ budget to handle finer details. If any inner refinement successfully proves at least one lemma (indicating meaningful progress), the outer refinement process will update this information into the prompt and continue its refinement.
Under this setting, Seed-Prover solves harder problems like IMO 2003 P6 ($\textit{MOHS}=35$), IMO 2020 P5 ($\textit{MOHS}=20$), IMO 2024 P1 ($\textit{MOHS}=5$) and IMO 2025 P5 ($\textit{MOHS}=15$), and final proofs potentially exceeding 1000 lines of code. 

\paragraph{Heavy} While the medium setting encourages reasoning in \textbf{depth} over proof details, it lacks the \textbf{breadth} needed to explore diverse properties of the given problem.
Under the heavy inference setting, Seed-Prover begins with a conjecture pool and an empty lemma pool for the given problem. Initially, the proposer generates thousands of conjectures (by default 5000) to populate the conjecture pool.
During inference, Seed-Prover tries to prove or disprove every conjecture in the conjecture pool using the light setting. Successfully proved conjectures are moved into the lemma pool. Seed-Prover leverages the lemma pool and Lean compiler feedback to refine the proof attempts.
Additional conjectures are proposed based on proved lemmas.
After days of thinking, the lemma pool accumulates several thousand nontrivial math facts. Each lemma is scored based on its proof rate, semantic relevance, and proof length. The proof rate serves as a strong indicator of lemma value; empirically, lemmas with low proof rate are often crucial to the final proof. Semantic relevance is assessed by an LLM judge. Unrelated lemmas or lemmas with short proof lengths are removed. We take hundreds of the top-ranked lemmas to help Seed-Prover finish the proof of the main problem using the medium inference setting. Seed-Prover has been trained to select and integrate these lemmas into a complete proof during RL.
IMO 2017 P2 ($\textit{MOHS}=40$), IMO 2025 P3 ($\textit{MOHS}=25$) and IMO 2025 P4 ($\textit{MOHS}=15$) are proved under the heavy inference setting. 

\section{Evaluation}

\subsection{Seed-Geometry}
Using Seed-Geometry's backend, we performed large-scale problem generation. In particular, we took a similar approach to TongGeometry \cite{zhang2024proposing} by collecting data statistics of geometry problems over more than the past 20 years, and ran the problem generation program on the trees. In more than 7 days, the problem generation program found over 230 million unique problems, reaching 8x the search efficiency compared to the Python implementation. After necessary preprocessing using a pretrained byte-pair encoding \cite{sennrich2015neural}, the dataset totaled 38B tokens. We trained both a policy model to complete the auxiliary objects given the context and a value model to estimate the number of steps remaining under the current state. The policy model was initialized from a pretrained Seed model and the value model was initialized with the trained policy model. However, in the experiments, we found that under extensive search, the value model could harm the general performance due to large errors in value estimation. We also compared generating auxiliary actions step-by-step with beam search and generating the whole auxiliary sequence in one go. Results from the latter were significantly inferior to the former. Therefore, in the final evaluation, we used step-by-step generation with beam search in a distributed setup, with each GPU process hosting a policy model for proposal generation.

\cref{tab:performance_imo} lists the performance of Seed-Geometry in IMO geometry problems from 2000 to 2024. Using the accounting method in IMO-AG-50, Seed-Geometry achieves 43 problem solves compared to AlphaGeometry 2, reaching 1 more solution. The problems that Seed-Geometry cannot solve but AlphaGeometry 2 does are not proof-based problems but rather computation-based problems, which AlphaGeometry 2 could potentially address using its algebraic engine.
\cref{tab:performance_imosl} lists the performance of our model in the much harder IMO shortlist problems from 2000 to 2022. The original benchmark of IMOSL-AG-30 was claimed to be established from the problems in the benchmark, but omitted many of the hard problems. Here, we include all hard-level problems in the shortlist and present the full results. As shown in the tables, AlphaGeometry 2 solved 19 of the 39 problems (with some unattempted) and our Seed-Geometry solved 22 of them.
In addition, Seed-Geometry solved IMO 2025 P2 in 2 seconds after it received the human-provided problem formulation.

Based on the results provided, Seed-Geometry has established a new state-of-the-art in automated geometry problem solving, surpassing the performance of its predecessor, AlphaGeometry 2. In summary, by solving more problems in both the standard IMO and the more difficult IMO shortlist benchmarks, Seed-Geometry demonstrates a superior capability in handling complex, proof-based geometry challenges, setting a new performance standard in the field.

\begin{table}[h!]
    \centering
    \caption{Performance comparison of AlphaGeometry 2 (AG2) and Seed-Geometry (SG) in IMO geometry problems from 2000 to 2024. Note that in IMO-AG-50, the 2002 P2, 2003 P4, 2004 P5, 2008 P1, and 2009 P4 are separated into two, whereas we merge each of them into one.}
    \label{tab:performance_imo}
    \vspace{1em} 
    \begin{minipage}[t]{0.32\textwidth} 
        \centering
        \begin{tabular}{>{\centering\arraybackslash}m{2cm} cc}
            \toprule
            \textbf{ID} & \textbf{AG2} & \textbf{SG} \\
            \midrule
            2000 P1  & \checkmark & \checkmark \\
            2000 P6  & \checkmark & \checkmark \\
            2001 P1  & \text{\sffamily X} & \text{\sffamily X} \\
            2001 P5  & \checkmark & \text{\sffamily X} \\
            2002 P2  & \checkmark & \checkmark \\
            2002 P6  & \text{\sffamily X} & \text{\sffamily X} \\
            2003 P3  & \text{\sffamily X} & \text{\sffamily X} \\
            2003 P4  & \checkmark & \checkmark \\
            2004 P1  & \checkmark & \checkmark \\
            2004 P5  & \checkmark & \checkmark \\
            2005 P1  & \checkmark & \checkmark \\
            2005 P5  & \checkmark & \checkmark \\
            2006 P1  & \text{\sffamily X} & \checkmark \\
            2006 P6  & \text{\sffamily X} & \text{\sffamily X} \\
            2007 P2  & \checkmark & \checkmark \\
            \bottomrule
        \end{tabular}
    \end{minipage}%
    \hfill
    \begin{minipage}[t]{0.32\textwidth} 
        \centering
        \begin{tabular}{>{\centering\arraybackslash}m{2cm} cc}
            \toprule
            \textbf{ID} & \textbf{AG2} & \textbf{SG} \\
            \midrule
            2007 P4  & \checkmark & \checkmark \\
            2008 P1  & \checkmark & \checkmark \\
            2008 P6  & \checkmark & \checkmark \\
            2009 P2  & \checkmark & \checkmark \\
            2009 P4  & \checkmark & \text{\sffamily X} \\
            2010 P2  & \checkmark & \checkmark \\
            2010 P4  & \checkmark & \checkmark \\
            2011 P6  & \checkmark & \checkmark \\
            2012 P1  & \checkmark & \checkmark \\
            2012 P5  & \checkmark & \checkmark \\
            2013 P3  & \checkmark & \checkmark \\
            2013 P4  & \checkmark & \checkmark \\
            2014 P3  & \checkmark & \checkmark \\
            2014 P4  & \checkmark & \checkmark \\
            2015 P3  & \checkmark & \checkmark \\
            \bottomrule
        \end{tabular}
    \end{minipage}%
    \hfill
    \begin{minipage}[t]{0.32\textwidth} 
        \centering
        \begin{tabular}{>{\centering\arraybackslash}m{2cm} cc}
            \toprule
            \textbf{ID} & \textbf{AG2} & \textbf{SG} \\
            \midrule
            2015 P4  & \checkmark & \checkmark \\
            2016 P1  & \checkmark & \checkmark \\
            2017 P4  & \checkmark & \checkmark \\
            2018 P1  & \checkmark & \checkmark \\
            2018 P6  & \text{\sffamily X} & \checkmark \\
            2019 P2  & \checkmark & \checkmark \\
            2019 P6  & \checkmark & \checkmark \\
            2020 P1  & \checkmark & \checkmark \\
            2020 P6  & \text{\sffamily X} & \text{\sffamily X} \\
            2021 P3  & \checkmark & \checkmark \\
            2021 P4  & \checkmark & \checkmark \\
            2022 P4  & \checkmark & \checkmark \\
            2023 P2  & \checkmark & \checkmark \\
            2023 P6  & \text{\sffamily X} & \checkmark \\
            2024 P4  & \checkmark & \checkmark \\
            \bottomrule
        \end{tabular}
    \end{minipage}
\end{table}

\begin{table}[h!]
    \centering
    \caption{Performance comparison of AlphaGeometry 2 (AG2) and Seed-Geometry (SG) in IMO shortlist geometry problems from 2000 to 2022. Note that in IMOSL-AG-30, many geometry problems are ignored and here we fill in those missing geometry problems for comparison, with 2016 G7 merged into one. NA represents ``not attempted'' as the problems results were not directly reported in the original paper.}
    \label{tab:performance_imosl}
    \vspace{1em} 
    \begin{minipage}[t]{0.32\textwidth} 
        \centering
        \begin{tabular}{>{\centering\arraybackslash}m{2cm} cc}
            \toprule
            \textbf{ID} & \textbf{AG2} & \textbf{SG} \\
            \midrule
            2002 G7  & \checkmark & \checkmark \\
            2002 G8  & \checkmark & \checkmark \\
            2003 G5  & \checkmark & \checkmark \\
            2004 G7  & \text{\sffamily X} & \checkmark \\
            2004 G8  & \textbf{NA} & \checkmark \\
            2005 G5  & \checkmark & \checkmark \\
            2005 G6  & \text{\sffamily X} & \checkmark \\
            2006 G9  & \checkmark & \checkmark \\
            2007 G8  & \text{\sffamily X} & \text{\sffamily X} \\
            2009 G6  & \checkmark & \checkmark \\
            2009 G7  & \checkmark & \text{\sffamily X} \\
            2009 G8  & \checkmark & \text{\sffamily X} \\
            2010 G5  & \checkmark & \checkmark \\
            \bottomrule
        \end{tabular}
    \end{minipage}%
    \hfill
    \begin{minipage}[t]{0.32\textwidth} 
        \centering
        \begin{tabular}{>{\centering\arraybackslash}m{2cm} cc}
            \toprule
            \textbf{ID} & \textbf{AG2} & \textbf{SG} \\
            \midrule
            2011 G3  & \text{\sffamily X} & \text{\sffamily X} \\
            2011 G4  & \textbf{NA} & \checkmark \\
            2011 G5  & \textbf{NA} & \checkmark \\
            2011 G6  & \checkmark & \text{\sffamily X} \\
            2011 G7  & \checkmark & \text{\sffamily X} \\
            2012 G6  & \text{\sffamily X} & \checkmark \\
            2012 G7  & \textbf{NA} & \text{\sffamily X} \\
            2012 G8  & \text{\sffamily X} & \checkmark \\
            2014 G7  & \checkmark & \checkmark \\
            2015 G5  & \checkmark & \checkmark \\
            2015 G7  & \textbf{NA} & \text{\sffamily X} \\
            2015 G8  & \textbf{NA} & \text{\sffamily X} \\
            2016 G5  & \checkmark & \checkmark \\
            \bottomrule
        \end{tabular}
    \end{minipage}%
    \hfill
    \begin{minipage}[t]{0.32\textwidth} 
        \centering
        \begin{tabular}{>{\centering\arraybackslash}m{2cm} cc}
            \toprule
            \textbf{ID} & \textbf{AG2} & \textbf{SG} \\
            \midrule
            2016 G6  & \checkmark & \checkmark \\
            2016 G7  & \checkmark & \checkmark \\
            2016 G8  & \textbf{NA} & \text{\sffamily X} \\
            2017 G7  & \text{\sffamily X} & \text{\sffamily X} \\
            2017 G8  & \textbf{NA} & \text{\sffamily X} \\
            2018 G7  & \checkmark & \checkmark \\
            2019 G6  & \checkmark & \checkmark \\
            2019 G8  & \textbf{NA} & \text{\sffamily X} \\
            2020 G8  & \checkmark & \checkmark \\
            2021 G8  & \text{\sffamily X} & \text{\sffamily X} \\
            2022 G6  & \text{\sffamily X} & \text{\sffamily X} \\
            2022 G7  & \text{\sffamily X} & \text{\sffamily X} \\
            2022 G8  & \textbf{NA} & \text{\sffamily X} \\
            \bottomrule
        \end{tabular}
    \end{minipage}
\end{table}

\subsection{Seed-Prover}
To evaluate the performance of Seed-Prover, we tested it on IMO 2025\footnote{ByteDance was officially invited to participate in IMO 2025.
}, past IMO problems, MiniF2F \citep{zhengminif2f}, PutnamBench \citep{tsoukalasputnambench}, CombiBench \citep{liu2025combibench}, and MiniCTX-v2 \citep{hu2025minictx} covering a wide range of mathematical domains.
For PutnamBench and CombiBench, we first evaluate using the light setting and use the medium setting for unsolved problems. For IMO problems and MiniF2F, we also use the heavy setting for the remaining unsolved problems. The results are shown in Table~\ref{tab:seed_prover_performance}. Unless otherwise specified, we use Lean v4.14.0 with its corresponding Mathlib version.

\begin{table}[t]
\centering
\caption{Performance comparison of Seed-Prover against previous systems across formal math tasks. The performance on PutnamBench is using the number of proved statements instead of percentage following previous works \citep{tsoukalasputnambench}.}
\begin{tabular}{lll}
\toprule
\textbf{Metric}               & \textbf{Seed-Prover}                              & \textbf{Previous SOTA}                     \\ \midrule
IMO 2025                      & 4/6 (Heavy, 5/6 post-competition)                       & 5/6 (Natural language, Gemini)\\
Past IMO                      & 78.1\% (Heavy)                                      & —                                          \\
MiniF2F-valid                 & 100.0\% (Medium\textsuperscript{1})              & 90.6\% (DeepSeek-Prover-V2 \citep{deepseek-v2})                \\
MiniF2F-test                  & 99.6\% (Medium\textsuperscript{2})                                     & 92.2\% (Kimina-Prover \citep{kimina})                     \\
PutnamBench                   & 331/657 (Medium)                                  & 86/657 (Goedel-Prover-V2 \citep{lin2025goedel})                  \\
CombiBench                    & 30.0\% (Medium)                                     & 10.0\% (Deepseek-Prover-V2 \citep{deepseek-v2})                  \\
MiniCTX-v2                    & 81.8\% (Light)                                    & 44.3\% (o4-mini \citep{minictx_leaderboard_website})                           \\ \bottomrule
\end{tabular}
{\\ \footnotesize
\textsuperscript{1}One problem used Heavy.
\textsuperscript{2}One problem failed under Heavy.}
\label{tab:seed_prover_performance}
\end{table}

\paragraph{IMO 2025} During the IMO 2025 contest, all problems were translated into formal statements by human experts. For fill-in-the-blank problems (``determine $x$ such that \dots''), initial solution candidates were generated by Seed1.6-Thinking before translation. We conducted searches for IMO 2025 Problems 1, 3, 4, and 5 using both the medium and heavy inference settings in parallel. As required by the IMO committee, all proof submissions were due by July 18th. Seed-Geometry solved Problem 2 instantly, and Seed-Prover derived the proof for Problem 5 under the medium inference setting, while proofs for the other three problems required the heavy inference setting. Notably, the proof for Problem 1 was finished after the deadline.

\paragraph{Past IMO} To evaluate our system's performance on past IMO problems, we curated a dataset consisting of 155 past IMO problems. Most problems were adapted from Compfiles\footnote{\url{https://github.com/dwrensha/compfiles}} and MiniF2F \citep{zhengminif2f}. Additionally, a subset of problems was manually added or corrected by human experts.
For problems prior to 2017, we used light and medium settings. For problems after 2017, we use the heavy inference setting if the medium inference setting failed.
Seed-Prover successfully proves 121/155 problems, achieving an overall success rate of $78.1\%$.
By difficulty, Seed-Prover proves 47/55 easy problems (P1 or P4), 47/56 medium problems (P2 or P5), and 27/44 hard problems (P3 or P6).
By subject area, it proves 72/85 algebra problems, 42/55 number theory problems, and 7/14 combinatorics problems.
This demonstrates that the Seed-Prover's performance at IMO 2025 reflects consistent capability on IMO problems across all years.

\paragraph{MiniF2F \citep{zhengminif2f}} Under the medium setting, we prove $99.6\%$ problems on both MiniF2F-valid and MiniF2F-test. We used the heavy inference setting to tackle the last problem in both splits (i.e. IMO 1990 P3 and IMOSL 2007 Algebra P6). Seed-Prover successfully proved IMO 1990 P3 and failed on IMOSL 2007 Algebra P6.
Interestingly, among the most challenging problems solved in MiniF2F are ones such as AMC12A 2021 P12, AMC12A 2021 P25, and AMC12A 2020 P9, which are relatively straightforward to reason about in natural language, but pose significant challenges when formalized in Lean. This difficulty arises primarily from obstacles in applying Vieta’s formulas or the non-triviality of counting roots.

\paragraph{PutnamBench \citep{tsoukalasputnambench}} We evaluated Seed-Prover on PutnamBench using the light and medium inference settings. Under the light inference setting only, Seed-Prover proved $201/657$ problems from PutnamBench. Using the medium inference setting improved this performance to $331/657$ problems. This result shows a significant performance jump compared to previous works on undergraduate math problems.

\paragraph{CombiBench \citep{liu2025combibench}} Previous benchmarks have primarily focused on number theory and algebra problems. CombiBench is a benchmark specifically centered on combinatorial problems, where the problems often involve newly-defined concepts. Here, we evaluate Seed-Prover on CombiBench using the medium inference setting. Our model proves 30 out of 100 problems from CombiBench, outperforming previous work. Nonetheless, relative to other benchmarks, our model still struggles with proving combinatorics problems.

\paragraph{MiniCTX-v2}\citep{hu2025minictx} To test our system on a broader range of mathematical subjects from real-world formalization projects—including the ability to understand new definitions, notations, and lemmas—we evaluated Seed-Prover on MiniCTX-v2. This dataset includes context-rich problems from formalization repositories in subjects such as axiomatic systems, high-energy physics, analysis, and research-level number theory, all of which were written after Nov.~2024 to prevent data contamination. For evaluation purposes, we used the light inference setting under Lean v4.16.0. Our system successfully achieved $81.8\%$ of MiniCTX-v2, which demonstrates its strong potential in real-world automated theorem proving, generalizing beyond standalone competition problems. For comparison, the baseline o4-mini solved $44.3\%$ statements at Pass@8 \citep{minictx_leaderboard_website}.
\section{Conclusion}
In this work, we presented Seed-Geometry and Seed-Prover---two formal reasoning frameworks that integrate the capabilities of large language models. Both systems substantially outperform previous formal reasoning frameworks. Seed-Geometry accelerates verification and scales the search mechanism. Seed-Prover leverages iterative refinement and a three-tiered test-time inference strategy to achieve state-of-the-art. Notably, we successfully proved 5 out of 6 problems in IMO 2025, demonstrating the efficacy of these formal systems. Formal languages like Lean offer rapid proof verification, making them far more cost-effective than human experts and more reliable than LLM judges. Our future work will focus on combining formal systems with large language models to tackle open conjectures.

\clearpage

\bibliographystyle{plainnat}
\bibliography{main}

\clearpage
\beginappendix
\section{Contributors}
The names are sorted alphabetically. An asterisk * indicates a member who left Seed.

\paragraph{Algorithm}
Luoxin Chen, Liankai Huang, Zhicheng Jiang, Allan Jie, Xiaoran Jin, Xing Jin, Chenggang Li, Wenlei Shi, Jiahui Wang, Siran Wang, Chenrui Wei, Shufa Wei, Yonghui Wu, Huajian Xin, Fan Yang, Hongyi Yuan, Zheng Yuan, Tianyang Zhan, Chi Zhang, Yue Zhang*, Yichi Zhou, Thomas Hanwen Zhu

\paragraph{Data}
Jinming Gu, Wenhao Huang, Zhicheng Jiang, Xiaoran Jin, Kaijing Ma, Jiawei Shen, Tong Sun, Chenrui Wei, Shufa Wei, Yuchen Wu, Yihang Xia, Huaiyuan Ying*, Zheng Yuan, Ge Zhang

\paragraph{Infra}
Cheng Ren, He Sun, Zhihong Wang, Tianyun Zhao*, Jianqiu Zhao, Thomas Hanwen Zhu

\section{LooKeng: An Easy-to-Use and Effective Python Interface for Lean}
\label{LooKeng}

Interacting with Lean poses significant challenges that limit the flexibility of Lean-based workflows. 
The most popular interface, LeanDojo \citep{yang2023Leandojo}, only supports earlier versions of Lean 4, restricting users from accessing Lean's newest updates.
Furthermore, LeanDojo requires creating a Lean repository for interaction, which makes it impractical to use considering the massive scale of Lean interaction during model development and inference.
To address these issues, we introduce LooKeng, a REPL\footnote{\href{https://github.com/leanprover-community/repl}{https://github.com/leanprover-community/repl}}-based Python interface designed to simplify and accelerate the interaction process. 
LooKeng offers powerful features for developers while providing a user-friendly interface for end-users.
The core functionality of LooKeng includes `init\_state', `run\_tac', and `verify\_proof'. One can use LooKeng to interact with Lean step-by-step or verify an entire proof directly.
The key features of LooKeng are summarized as follows:
\begin{itemize}
    \item \textbf{Stateless Design}: A Lean state can be simultaneously processed using different LooKeng instances, enabling effortless scaling and sharing.
    
    \item \textbf{Complex Tactics}: Complex tactics such as \texttt{apply?} and \texttt{all\_goals} are fully supported, with enhanced infotree integration to prevent false positive proofs.

    \item \textbf{Version-Free}: The LooKeng CLI allows users to manage and switch between different Lean versions. 
    
    \item \textbf{Memory Control}: Users can easily track the memory consumption of the Lean backend, set custom thresholds, and automatically terminate processes when memory usage exceeds the limit.

     \item \textbf{Proof Verification}: LooKeng provides a straightforward method, `verify\_proof', to rigorously verify the final proof using the native Lean interface, ensuring correctness and reliability.

     \item \textbf{Proof Simplification}: LooKeng can remove useless tactics and hypothesis in the proof to obtain a simpler proof.

     \item \textbf{Statement Negation}: LooKeng is able to generate the negated statement of a statement.

     \item \textbf{Multi-Concurrency Support}: LooKeng can run as a service, handling thousands of concurrent requests via async architecture and resource isolation.

\end{itemize}


\end{document}